\documentclass[final]{elsarticle}

\usepackage{lineno,hyperref}

\usepackage{url}

\usepackage{amsmath,amsfonts,bbm}
\usepackage{bbding}
\usepackage{algorithmic}
\usepackage{graphicx}
\usepackage{textcomp}
\usepackage{overpic}
\usepackage{subfigure}
\usepackage{threeparttable}
\usepackage{makecell}
\usepackage{multirow}
\usepackage{booktabs}
\usepackage{cases}
\usepackage{geometry}
\modulolinenumbers[5]

\journal{Ultrasonics}

\begin{document}

\begin{frontmatter}

\title{When SAM Meets Sonar Images}

\author[1,2]{Lin Wang} 
\author[2]{Xiufen Ye\corref{cor}}
\author[1]{Liqiang Zhu}
\author[1]{Weijie Wu}
\author[3]{Jianguo Zhang}
\author[2]{Huiming Xing}
\author[1]{Chao Hu\corref{cor}}

\cortext[cor]{Corresponding author. \\ 
\indent\; E-mail: huchao.000@gmail.com (Chao Hu) \\ 
\indent\qquad\qquad\ yexiufen@hrbeu.edu.cn (Xiufen Ye)
}

\address[1]{China Unicom (Shanghai) Industry Internet Co., Ltd., Shanghai, China.}
\address[2]{College of Intelligent Systems Science and Engineering, Harbin Engineering University, Harbin, China}
\address[3]{Shanghai Graphic Design Information Co., Ltd, Shanghai, China}

\begin{abstract}
Segment Anything Model (SAM) has revolutionized the way of segmentation. However, SAM's performance may decline when applied to tasks involving domains that differ from natural images. Nonetheless, by employing fine-tuning techniques, SAM exhibits promising capabilities in specific domains, such as medicine and planetary science. Notably, there is a lack of research on the application of SAM to sonar imaging. In this paper, we aim to address this gap by conducting a comprehensive investigation of SAM's performance on sonar images. Specifically, we evaluate SAM using various settings on sonar images. Additionally, we fine-tune SAM using effective methods both with prompts and for semantic segmentation, thereby expanding its applicability to tasks requiring automated segmentation. Experimental results demonstrate a significant improvement in the performance of the fine-tuned SAM. The code will be available at \url{https://github.com/wangsssky/SonarSAM}.
\end{abstract}


\end{frontmatter}

\section{Introduction}

The propagation of light in underwater environments is impeded by the high absorption and scattering of light waves~\cite{shen2021underwater}, which restricts the effective range of optical-based visual detection to short distances. This limitation necessitates the use of alternative sensing modalities, such as acoustic imaging, which offers a long detection range due to the negligible effects of water and dissolved impurities on the propagation of sound waves. This feature is particularly advantageous for a range of marine applications, including scientific research, fishery, and military. 

Image segmentation is the process of dividing an image into multiple regions, each corresponding to a different object. Segmentation generates a fine description of the targets than the bounding boxes or types provided by detection and classification. However, sonar image segmentation can be a challenging task due to the high noise level of the image, low resolution, and complex object shapes. 

Since AlexNet~\cite{krizhevsky2017imagenet} was proposed to boost the classification performance with a large margin on ImageNet~\cite{deng2009imagenet} in 2012, deep-learning techniques have undergone significant development and have achieved encouraging results in various domains, including natural language processing, computer vision, speech recognition, and so on. In particular, deep learning has also shown promising results in improving sonar image segmentation. 
However, despite these advancements, the field of sonar imaging has often lagged several years behind cutting-edge technologies~\cite{steiniger2022survey}.

The Segment Anything Model (SAM)~\cite{kirillov2023segment} has been developed as a general segmentation model that has been trained on the largest dataset to date for segmentation, comprising 11 million images and 1 billion masks. The exceptional performance of SAM across numerous segmentation tasks has led some researchers to believe that the image segmentation task has been resolved. However, subsequent research has shown that SAM's performance may exhibit significant degradation for specific scenarios, especially the images with irregular shapes, weak boundaries, small size, and low contrast in different imaging modalities~\cite{zhang2023how}, such as medical images\cite{he2023accuracy,huang2023segment,mazurowski2023segment} and camouflage images~\cite{chen2023sam}, which may be due to the absence of the corresponding image samples during SAM training. To address these issues, several fine-tuning methods have been proposed to improve the performance of SAM on specific tasks~\cite{MedSAM,wu2023medical,qiu2023learnable,li2023polyp,zhang2023customized,julka2023knowledge}.

However, to the best of our knowledge, there is no research regarding the performance of SAM on sonar images, and as such, there is a lack of information on potential fine-tuning methods that may be required to adapt the model to this domain.

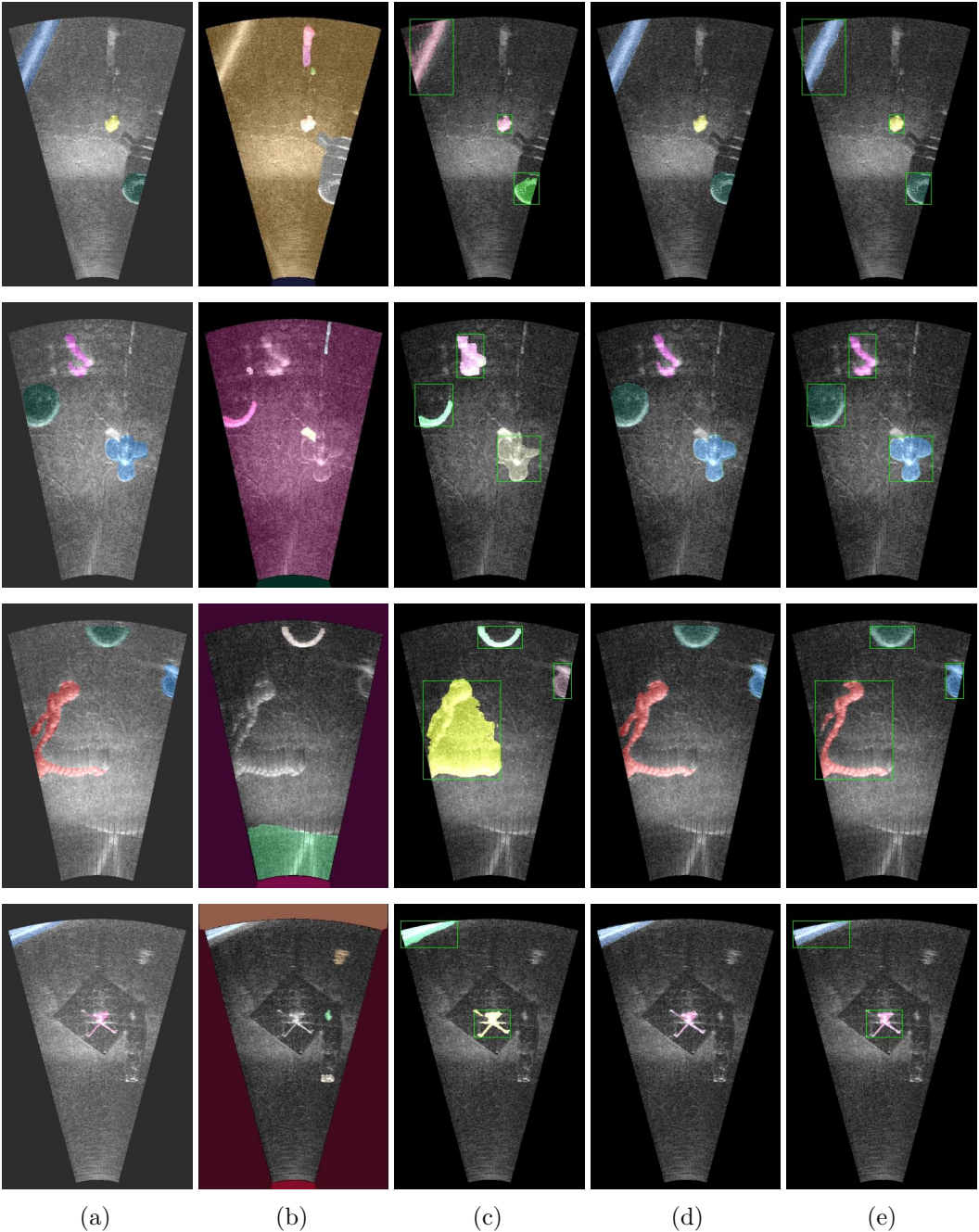
\begin{figure}[!t]
	\begin{center}
		\begin{overpic}[width=\columnwidth]{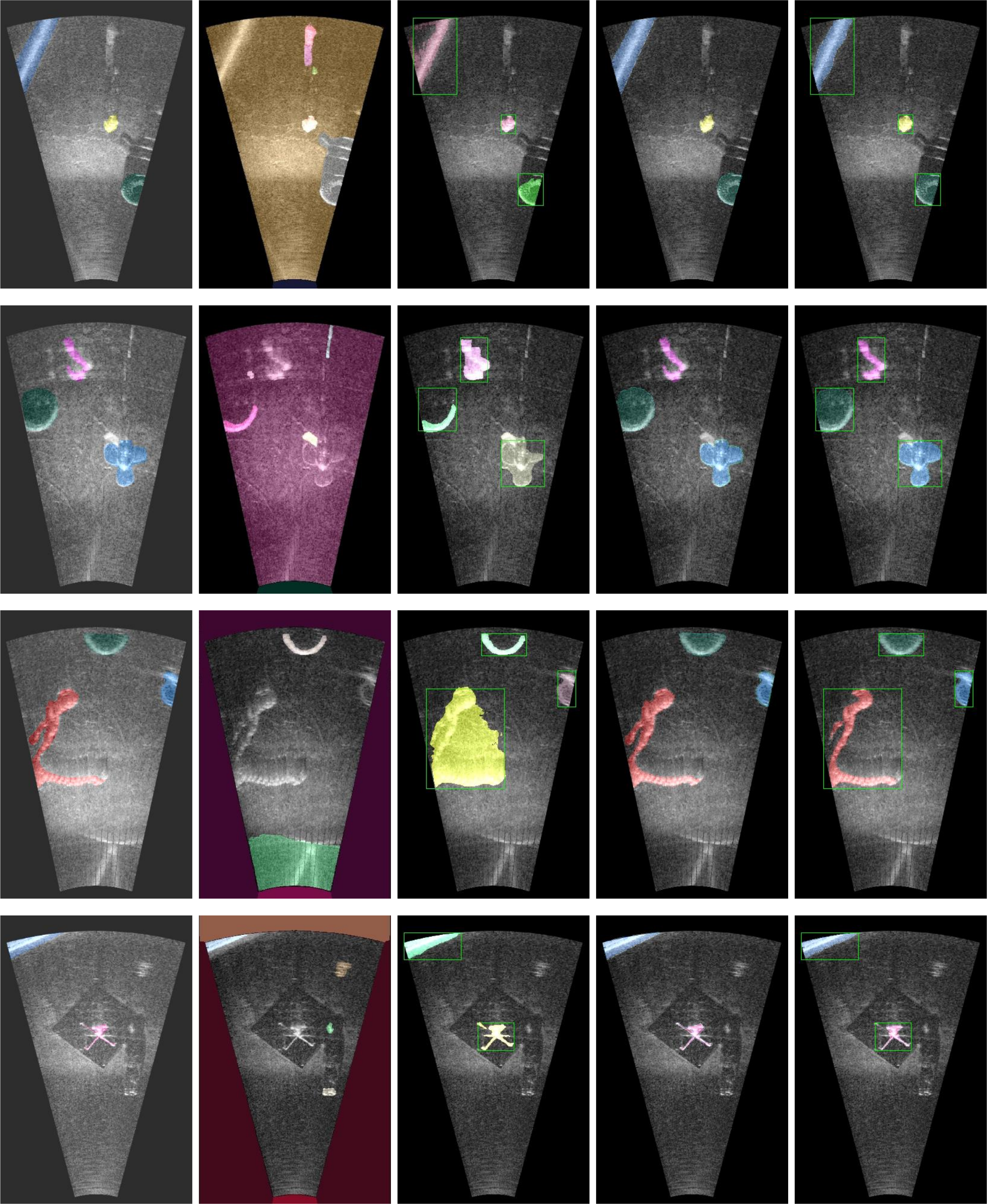}
			\put(6.5,-3){(a)}
			\put(23,-3){(b)}			
			\put(39.6,-3){(c)}
			\put(56.3,-3){(d)}
			\put(72.8,-3){(e)}
		\end{overpic}
	\end{center}
	\caption{
	Results of applying SAM to forward-looking sonar images. From left to right: (a) the ground truth; (b) results of the original SAM with default settings; (c) results of the original SAM with box prompts; (d) results of the fine-tuned SAM for semantic segmentation task; (e) results of the fine-tuned SAM with box prompts. 
	}
	\label{fig:visual_results}
\end{figure}

This paper aims to address the aforementioned gap in research by introducing SAM to sonar images and evaluating its performance. Furthermore, we conduct a comprehensive investigation of mainstream fine-tuning methods for SAM on sonar images, both qualitatively and quantitatively. We hope this work can help bridge the gap between large vision models and their applications in sonar images, thus promoting further research in this field. 

We specifically evaluate the performance of SAM on forward-looking sonar images. Similar to other domains that differ from natural images, we observe a significant decrease in results, as depicted in Figure~\ref{fig:visual_results} (b) and (c). Hence, there is a necessity to fine-tune SAM for sonar images to enhance its performance. We select recent promising fine-tuning methods for SAM and conduct a comprehensive evaluation on sonar images. Furthermore, to address the requirement for automatic segmentation without human interaction, we explore fine-tuning SAM for semantic segmentation in addition to generating results with prompts.

The contributions of the paper are concluded as follows:
\begin{enumerate}
    \item To the best of our knowledge, this study presents the first investigation of SAM applied to sonar images, thereby benefiting the sonar imaging community and expanding the application of SAM.
	\item Comprehensive fine-tuning methods for SAM are employed and assessed on sonar images, providing an experience for fine-tuning SAM on specific tasks with limited data samples and domain gaps compared to natural images.
	\item In addition to the general segmentation with prompts, we apply the semantic segmentation fine-tuning for SAM, enabling automatic segmentation without the need for additional interactions.
\end{enumerate}

\section{Related works}
\subsection{Sonar image segmentation}

As a subfield of image segmentation, segmentation methods for sonar images generally follow the same path as those for natural images, with popular models such as UNet~\cite{ronneberger2015u}, FCN~\cite{long2015fully}, SegNet~\cite{badrinarayanan2017segnet}, and DeeplabV3~\cite{chen2018encoder} being commonly used. 
For instance, Zheng et al.~\cite{zheng2021universal} introduced DeeplabV3+ to distinguish the water column from the seafloor. Song et al.~\cite{8232312} applied FCN on sonar images and segmented the sonar images into highlight, shadow, and background regions. Rahnemoonfar and Dobbs~\cite{rahnemoonfar2019semantic} used a UNet-like structure to segment the potholes in a seagrass bed.

\subsection{Segment anything model}

\begin{figure}[!t]
	\begin{center}
		\begin{overpic}[width=0.8\columnwidth]{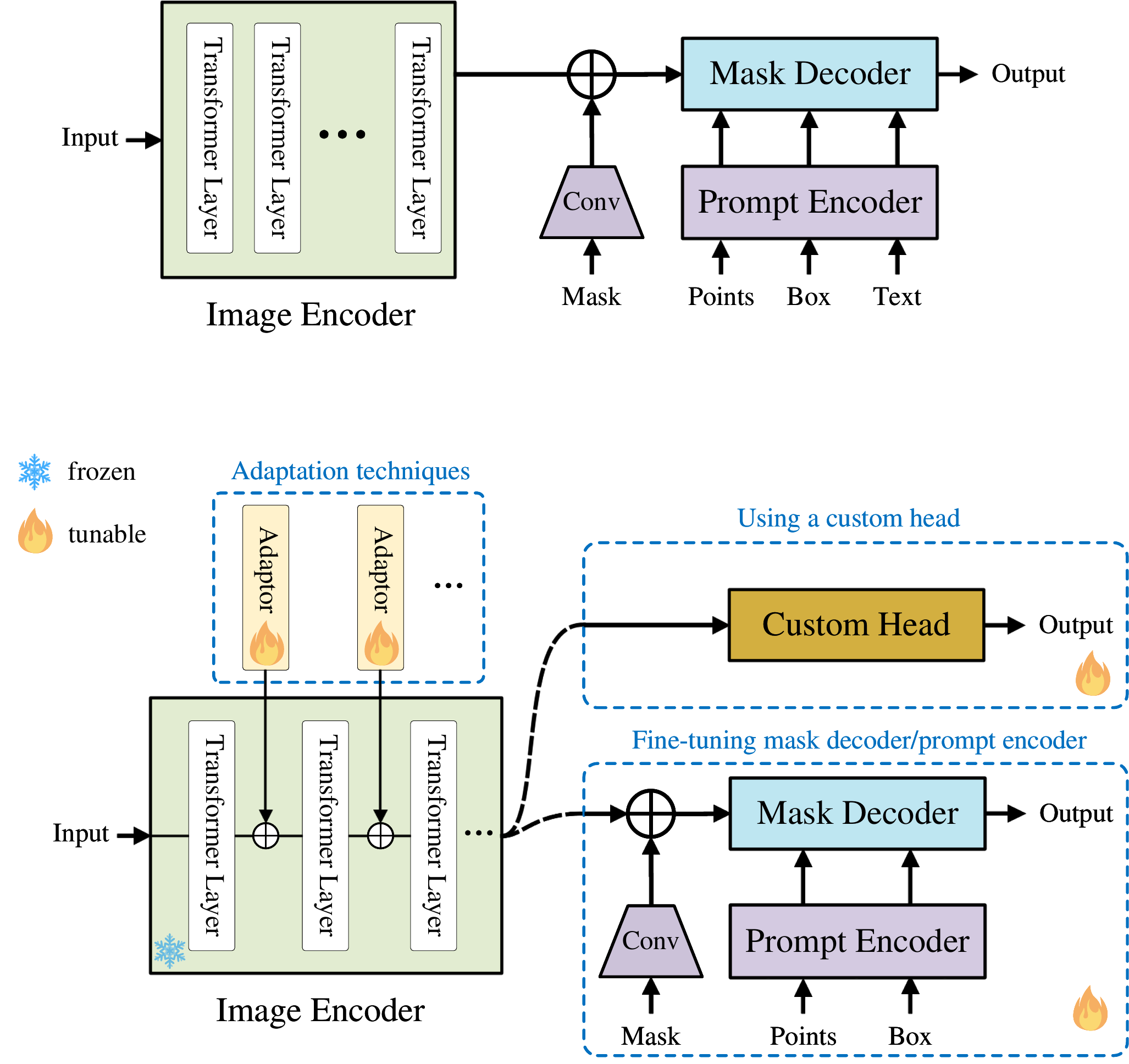}
			\put(49,60){(a)}
			\put(49,-3){(b)}			
		\end{overpic}
	\end{center}
	\caption{
	Illustration of the SAM network structure (a) and the mainstream methods for adapting SAM for specific tasks (b), i.e., fine-tuning the image encoder with adaptation methods, fine-tuning the light-weight mask decoder and prompt encoder, and using a custom segmentation head.
	}
	\label{fig:fine_tune_types}
\end{figure}

SAM~\cite{kirillov2023segment} is a state-of-the-art general image segmentation model that has been trained on a large natural image dataset. This model can segment objects using various prompts, such as points, boxes, masks, and text, and it exhibits excellent zero-shot learning capability, making it popular in a wide range of imaging applications. The SAM architecture consists of three main components: an image encoder, a prompt encoder, and a mask decoder. The image encoder utilizes a Vision Transformer (ViT)~\cite{dosovitskiy2010image}-based feature extractor to produce image embeddings while the prompt encoder processes the segmentation prompts. Finally, the mask decoder generates the segmentation masks. Figure~\ref{fig:fine_tune_types} (a) provides an illustration of the SAM architecture.

While SAM has demonstrated strong generalization capabilities, it has been observed to exhibit degraded performance on certain tasks. To address this issue, several fine-tuning methods have been proposed.
Ma et al.~\cite{MedSAM} proposed MedSAM as a method to adapt SAM for medical images. MedSAM fine-tunes only the mask decoder with bounding boxes as the prompts while keeping the other parts of the network frozen. Besides fine-tuning the mask decoder only, Li et al.~\cite{li2023polyp} fine-tuned the entire network and achieved slightly better performance than fine-tuning the mask decoder only, with a marginal improvement (less than 3\% in DICE score).

Zhang et al.~\cite{zhang2023customized} proposed SAMed, which fine-tuned only a small fraction of the parameters by using the low-rank adaptation (LoRA)~\cite{hu2021lora}. Additionally, they modified the mask decoder to provide semantic information. Wu et al.~\cite{wu2023medical} introduced an adaptation technique~\cite{chen2022adaptformer} to fine-tune SAM for medical imaging scenarios. Chen et al.~\cite{chen2023sam} also proposed SAM-Adapter to address underperformance scenarios such as camouflage and shadow.

Qiu et al.~\cite{qiu2023learnable} presented a visual prompt tuning method specifically for ophthalmology images. They froze the weights of the transformer layers in the ViT backbone and added tunable convolution layers between the transformer layers to learn task-specific information. The prompt encoder and mask decoder were replaced with a custom head. 
Julka and Granitzer~\cite{julka2023knowledge} proposed the application of SAM for landform segmentation in planetary science. However, they observed that while fine-tuning the mask decoder of SAM resulted in some spurious results, the optimal performance could be achieved with additional prompts, making it impractical for autonomous deployment without human involvement. To address this, they employed a lightweight custom decoder while using the SAM decoder to label incremental samples. 

In conclusion, the fine-tuning methods for SAM primarily focus on three aspects: (1) fine-tuning the image encoder with adaptation techniques; (2) fine-tuning the light-weight prompt encoder and/or mask decoder; and (3) using a custom head for specific tasks, as illustrated in Figure~\ref{fig:fine_tune_types} (b). 
However, the performance of these models on sonar images and the optimal approach for sonar image segmentation tasks remain unknown. Hence, our work conducts extensive experiments on SAM with sonar images, which can contribute to future research on fine-tuning large vision models for specific image domains.

\section{Method}

\begin{figure}[!t]
	\begin{center}
		\begin{overpic}[width=\columnwidth]{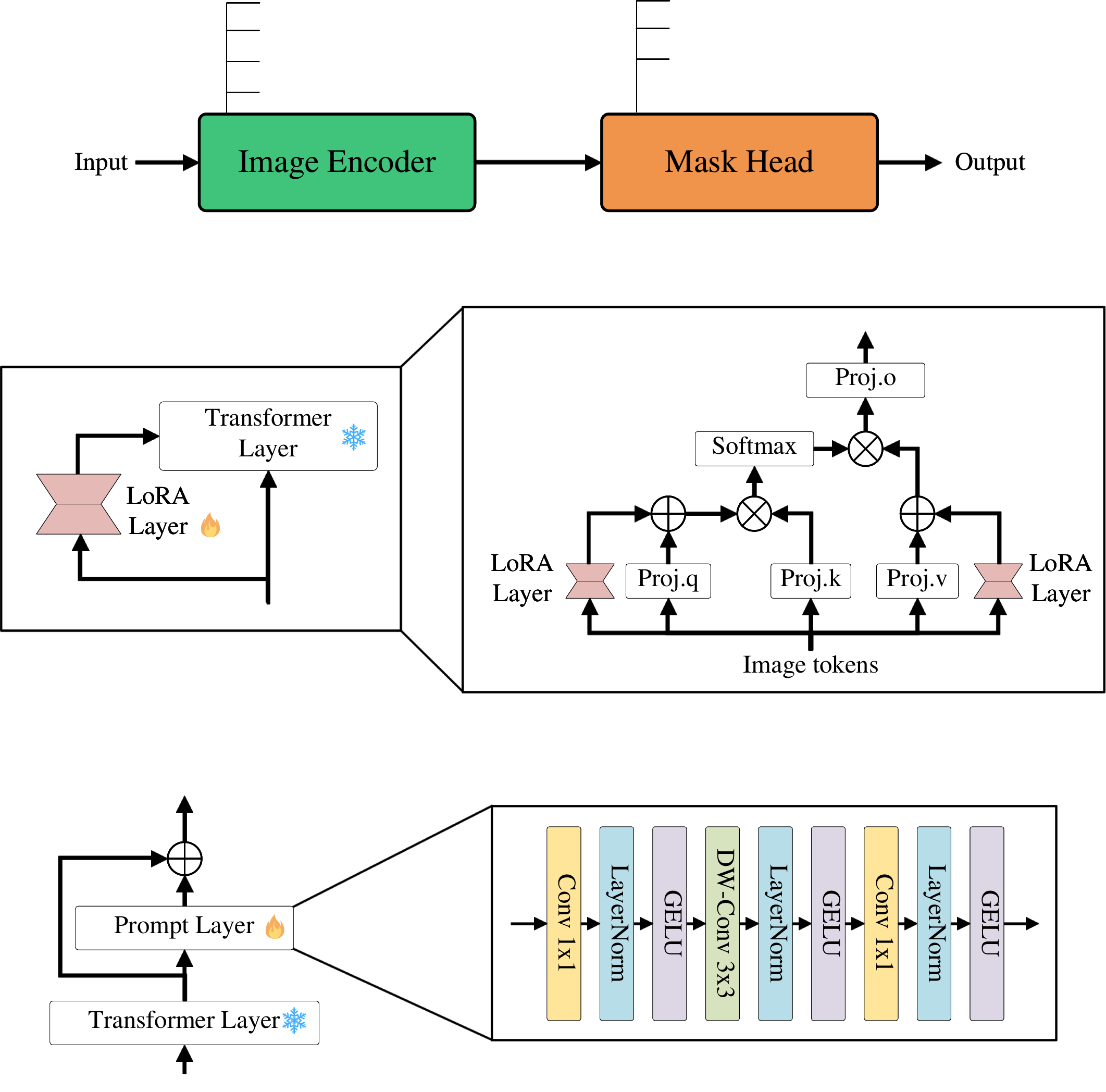}
			\put(40,75){(a) Overview}
			\put(28,30){(b) Fine-tuning with LoRA~\cite{zhang2023customized}}			
			\put(26,-3){(c) Fine-tuning with Prompt layer~\cite{qiu2023learnable}}
			\put(25,97.5){\scriptsize{Frozen}}
			\put(25,95){\scriptsize{Full-tuning}}
			\put(25,92.5){\scriptsize{Fine-tuning with LoRA}}
			\put(25,90){\scriptsize{Fine-tuning with Prompt layer}}
			\put(62,97.5){\scriptsize{Full-tuning}}
			\put(62,95){\scriptsize{Fine-tuning with LoRA}}
			\put(62,92.5){\scriptsize{Custom segmentation head}}
		\end{overpic}
	\end{center}
	\caption{
	The fine-tuning settings and the major structures of the adaptation methods for SAM. 
	}
	\label{fig:structure}
\end{figure}
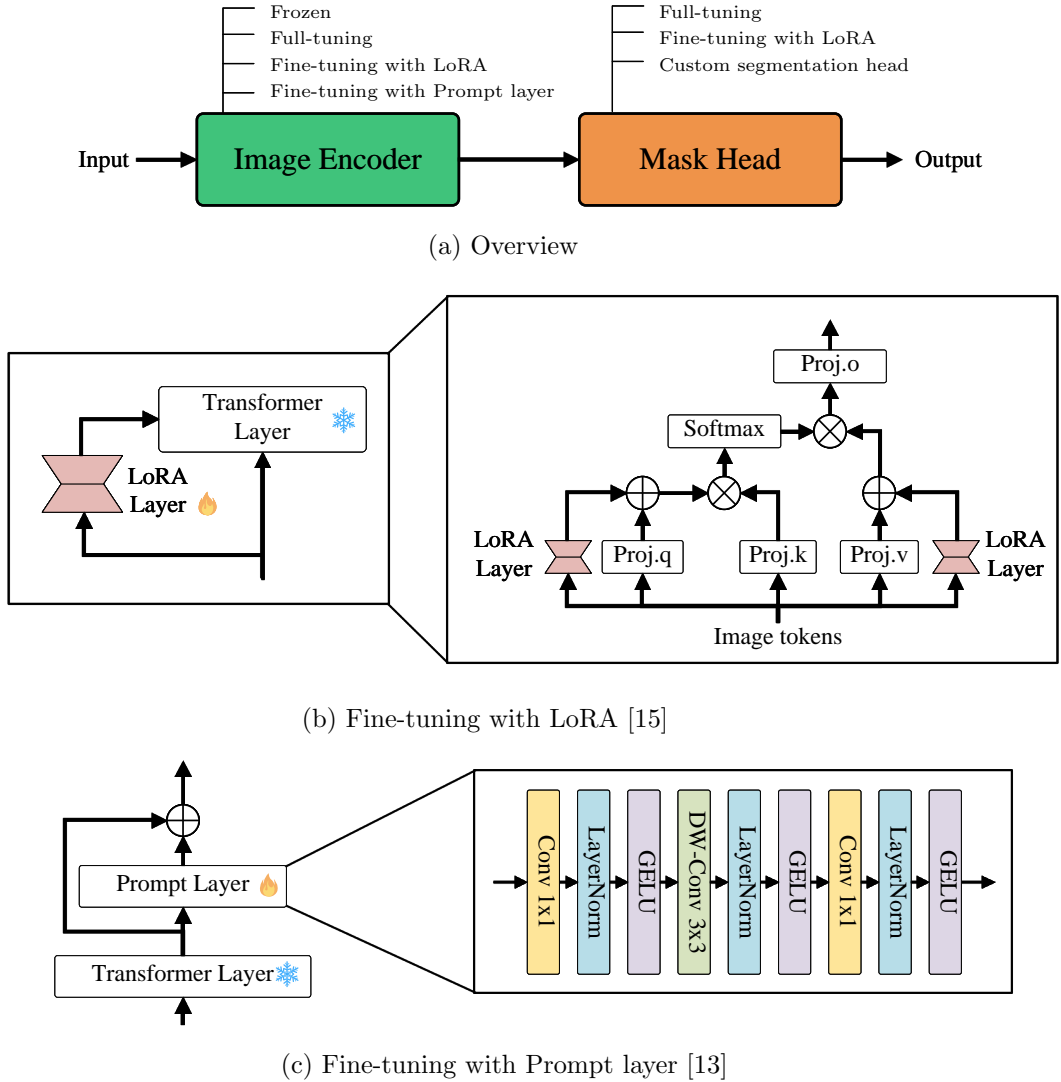

In this section, we present our framework for fine-tuning SAM. We have carefully selected a set of promising fine-tuning methods that are currently being applied to the Image encoder and Mask decoder components of SAM. 
Specifically, for the Image encoder, we evaluate four fine-tuning settings: frozen, full-tuning, fine-tuning with LoRA, and fine-tuning with prompt layer. Regarding the Mask decoder, we employ three settings: full-tuning, fine-tuning with LoRA, and custom segmentation head. 
To provide a visual representation of these fine-tuning methods, Figure~\ref{fig:structure} illustrates their corresponding structures.

\subsection{Fine-tuning settings for Image encoder}\label{fine_tuning_for_Image_encoder}
Four fine-tuning settings are employed for the Image encoder, and they are described as follows.

\noindent \textit{Frozen:} 
During training, a commonly employed transfer learning strategy involves freezing the parameters of the Image encoder and solely focusing on fine-tuning the Prompt encoder and the Mask decoder~\cite{MedSAM}. This approach is effective in reducing computation requirements, as the Image encoder accounts for the majority of SAM's parameters.

\noindent \textit{Full-tuning:} In contrast to freezing the Image encoder, fully fine-tuning involves updating all the parameters of the Image encoder~\cite{li2023polyp}. Fully fine-tuning typically demands significantly higher computational resources and necessitates a longer training duration.

\noindent \textit{Fine-tuning with LoRA:} In contrast to fully tuning all parameters in the Image encoder, LoRA enables model updates by tuning only a small number of the parameters. Figure~\ref{fig:structure} (b) illustrates the application of LoRA to the transformer layer in the Image encoder~\cite{zhang2023customized}. The LoRA layer comprises two linear layers with a small latent dimension (rank), such as 4, as employed in this study.

\noindent \textit{Fine-tuning with Prompt layer:} Fine-tuning with Prompt layer introduce visual prompt tuning to SAM~\cite{qiu2023learnable,jia2022vpt}. Similar to tuning the Image encoder with LoRA, only a limited number of parameters undergo tuning. Figure~\ref{fig:structure} (c) illustrates the structure of the Prompt layer.

\subsection{Fine-tuning settings for Mask decoder}
Three settings are employed for the Mask decoder: full-tuning, fine-tuning with LoRA, and custom segmentation head. The idea of the first two settings is introduced in section~\ref{fine_tuning_for_Image_encoder}. 

\noindent \textit{Custom Segmentation head:} The lack of semantic labeling in the results of SAM poses challenges for tasks relying on class information. To address this, the most direct way is to integrate an Image encoder with a customized segmentation head~\cite{julka2023knowledge,zhang2023customized,qiu2023learnable}. Our approach builds upon the simple segmentation head introduced in~\cite{qiu2023learnable,zhang2022hard}, comprising transpose convolutional layers, grouped convolutional layers, and a linear layer.

\subsection{Loss functions}
We use a joint loss of cross-entropy loss and DICE loss with \textit{Laplace smoothing} to guide the training of the segmentation task, which is defined as follows:
\begin{equation}
	\mathcal{L}_\mathit{CE} = -\frac{1}{N}\sum_{i=1}^{N}\sum_{j=1}^{C} y_{ij} \log(p_{ij}),
	\label{eq:CE_loss}
\end{equation}
where $N$ represents the total number of pixels in an image, $C$ represents the number of classes, $y_ {ij}$ is the ground truth label for the $i$-th pixel for the $j$-th class. It is $1$ if the sample belongs to the class and $0$ otherwise. $p_{ij}$ is the predicted probability of the $i$-th sample belonging to the $j$-th class.
\begin{equation}
	\mathcal{L}_\mathit{DICE} = 1 - \frac{2|\widehat{Y}\cap Y|+1}{|\widehat{Y}|+|Y|+1},
	\label{eq:DICE_loss}
\end{equation}
where $\widehat{Y}$ and $Y$ are the predicted and ground truth masks, respectively.
\begin{equation}
	\mathcal{L}_\mathit{joint} = \mathcal{L}_\mathit{CE} + \mathcal{L}_\mathit{DICE}.
	\label{eq:joint_loss}
\end{equation}

\section{Experiments and results}
To comprehensively evaluate the ability of SAM in sonar image scenarios, we conduct experiments on three aspects: 1) Evaluating the original SAM on sonar images; 2) Fine-tuning the SAM with various settings on sonar images; 3) Fine-tuning and adapting SAM on sonar images for task-specific semantic segmentation.

\subsection{Dataset}
We conduct the experiments on the Marine Debris dataset~\cite{singh2021marine}, which is the largest public-accessible real-world forward-looking sonar for semantic segmentation currently. The Marine Debris dataset contains 1868 images captured by ARIS Explorer 3000 sensor. Eleven types of objects were labeled, including bottle, can, chain, drink-carton, hook, propeller, shampoo-bottle, standing-bottle, tire, valve, and wall. 
As there is no official division of the dataset, we split the dataset into training, validation, and testing sets randomly with a ratio of 6: 2: 2. 

\subsection{Implementation details}
In order to keep consistent with the input of SAM, we convert the gray-scale sonar images into RGB color space and scale the image size to $1024\times1024$ during preprocessing. 
We introduce random flipping and color jittering as augmentation.
The batch size is set to $4$. The ADAM optimizer~\cite{kingma2014adam} is deployed for training.
The base learning rate is set to $3\times 10^{-4}$, and the cosine decay learning rate scheduler~\cite{LoshchilovH17} is used with a 1-epoch warm-up. All the models are trained for 30 epochs on the training set, and the hyperparameters are tuned on the validation set. The final performance of the models is evaluated on the testing set. 
The codes are implemented in PyTorch, and the experiments are conducted with two NVIDIA A100 GPUs. 

\subsection{Evaluating SAM on sonar images}
In this section, we first apply the original SAM for the sonar image segmentation task. We compare the SAM (vit-h version) with various popular segmentation models, including UNet~\cite{ronneberger2015u}, SegResNet~\cite{myronenko20193d}, FCN~\cite{long2015fully}, and DeeplabV3~\cite{chen2018encoder}, on Marine Debris dataset. We evaluate SAM with two prompt settings: 1) the default setting that uses the grid points as input; 2) using the box prompts that inputs the bounding boxes of the targets. 
The quantitive performance comparisons are shown in Table~\ref{tab:naive_SAM_results}. 

\begin{table}[tbph]
    \footnotesize        
        \caption{The performance of SAM on sonar images evaluated by DICE score.}
        \centering
		\setlength\tabcolsep{4.5pt}
        \begin{threeparttable}		
			\begin{tabular}{ccccccccccccc}			
			\toprule[1.5pt]			
			\multicolumn{1}{c}{\multirow{2}{*}{\textbf{Model}}} & \multicolumn{12}{c}{\textbf{DICE Score $(\%)$}} \\ 			
			\cline{2-13} 
			\specialrule{0em}{2pt}{2pt}
			\multicolumn{1}{c}{}                       & \multicolumn{1}{c}{\rotatebox[origin=l]{90}{bottle}} & \multicolumn{1}{c}{\rotatebox[origin=l]{90}{can}} & \multicolumn{1}{c}{\rotatebox[origin=l]{90}{chain}} & \multicolumn{1}{c}{\rotatebox[origin=l]{90}{\makecell[l]{drink\\carton}}} & \multicolumn{1}{c}{\rotatebox[origin=l]{90}{hook}} & \multicolumn{1}{c}{\rotatebox[origin=l]{90}{propeller}} & \multicolumn{1}{c}{\rotatebox[origin=l]{90}{\makecell[l]{shampoo\\bottle}}} & \multicolumn{1}{c}{\rotatebox[origin=l]{90}{\makecell[l]{standing\\bottle}}} & \multicolumn{1}{c}{\rotatebox[origin=l]{90}{tire}} & \multicolumn{1}{c}{\rotatebox[origin=l]{90}{valve}} & \multicolumn{1}{c}{\rotatebox[origin=l]{90}{wall}} & \multicolumn{1}{c}{\rotatebox[origin=l]{90}{\textbf{average}}} \\ 
			\midrule[0.8pt]
			UNet~\cite{ronneberger2015u}&	
			67.35&	55.64&	25.79&	8.10&	0.02&	0.23&	65.24&	1.87&	75.40&	37.82&	86.76&	38.57\\
			SegResNet~\cite{myronenko20193d}&	
			74.81&	67.96&	55.69&	5.35&	0.14&	26.58&	75.79&	0.03&	83.73&	49.33&	88.55&	48.00\\
			FCN~\cite{long2015fully}&	
			82.58&	77.17&	72.77&	83.36&	65.58&	72.06&	73.36&	83.62&	86.92&	66.60&	87.00&	77.37\\
			FCN$^*$&	
			82.51&	78.60&	77.61&	86.17&	76.44&	83.83&	79.20&	89.92&	89.87&	71.13&	87.79&	82.10\\
			DeeplabV3~\cite{chen2018encoder}&	
			83.88&	77.57&	73.97&	83.11&	73.93&	80.67&	81.98&	90.29&	85.37&	64.76&	88.13&	80.33\\
			DeeplabV3$^*$&	
			86.43&	77.38&	75.36&	85.52&	76.48&	83.64&	82.20&	90.24&	86.85&	65.46&	87.31&	81.53\\
			\midrule[0.8pt]
			\makecell{SAM\\(default)}&	
			5.82&	7.80&	1.44&	19.98&	5.84&	5.47&	9.21&	3.76&	7.57&	4.97&	7.75&	7.24\\
			\makecell{SAM\\(box prompt)}&	
			63.25&	32.04&	35.42&	13.72&	9.74&	41.26&	70.34&	60.35&	33.35&	36.52&	46.99&	40.27\\
			\bottomrule[1.5pt]
			\end{tabular}
        \begin{tablenotes}
            \item The average DICE score is calculated by averaging the DICE scores of each class. $^*$ indicates the weights initialized with a pre-trained model trained on COCO dataset~\cite{cocodataset}. 
        \end{tablenotes}
        \end{threeparttable}
        \label{tab:naive_SAM_results}
    \end{table}

Table~\ref{tab:naive_SAM_results} demonstrated that SAM has a significant performance drop when applied to the sonar images, especially for the default setting, only achieving a 7.24\% average DICE score. This may be due to the large domain gap between the source images for training SAM and sonar images. The former images are mainly photographs with high quality\cite{kirillov2023segment}, while the images of forward-looking sonar are in grayscale with low resolution and high noises. With the location information provided by the box prompts, the performance of the DICE score witnesses an obvious increase, reported as 40.27\%, comparable to the performance of UNet. However, there is still a large margin to the best performance achieved by the FCN~\cite{long2015fully}.
Figure~\ref{fig:visual_results} (b) and (c) depict the outcomes of applying SAM directly to sonar images. The segmentation results exhibit limited success in accurately delineating the targets, particularly in areas characterized by intricate structures and non-uniform intensities.

\subsection{Fine-tuning SAM for sonar images}\label{Fine_tuning_SAM}

In this section, we address the issue of a significant performance drop observed when applying SAM directly to sonar images. To mitigate this, we propose and evaluate several fine-tuning methods for model optimization. Specifically, we investigate three fine-tuning approaches for the Image encoder: freezing the Image encoder, fine-tuning the Image encoder with LoRA, and fine-tuning the Image encoder with the prompt layer. Additionally, we explore two methods for fine-tuning the Mask decoder: using LoRA and full-tuning. 
We use the vit-l backbone for the Image encoder. During the training process, only the box prompts are utilized.
The experimental results, presented in Table~\ref{tab:finetune_SAM_results}, provide insights into the effectiveness of these fine-tuning strategies.

\begin{table}[tbph]
    \footnotesize        
        \caption{The performance of SAM on sonar images evaluated by DICE score.}
        \centering
		\setlength\tabcolsep{4.5pt}
        \begin{threeparttable}		
			\begin{tabular}{cccccccccccccc}			
			\toprule[1.5pt]	
			\multicolumn{1}{c}{\multirow{2}{*}{\rotatebox[origin=l]{90}{\textbf{\makecell[l]{Image\\encoder\quad\quad}}}}} & 
			\multicolumn{1}{c}{\multirow{2}{*}{\rotatebox[origin=l]{90}{\textbf{\makecell[l]{Mask\\decoder\quad\quad}}}}} & 
			\multicolumn{12}{c}{\textbf{DICE Score $(\%)$}} \\ 			
			\cline{3-14} 
			\specialrule{0em}{2pt}{2pt}
			\multicolumn{1}{c}{}	&\multicolumn{1}{c}{}                       & \multicolumn{1}{c}{\rotatebox[origin=l]{90}{bottle}} & \multicolumn{1}{c}{\rotatebox[origin=l]{90}{can}} & \multicolumn{1}{c}{\rotatebox[origin=l]{90}{chain}} & \multicolumn{1}{c}{\rotatebox[origin=l]{90}{\makecell[l]{drink\\carton}}} & \multicolumn{1}{c}{\rotatebox[origin=l]{90}{hook}} & \multicolumn{1}{c}{\rotatebox[origin=l]{90}{propeller}} & \multicolumn{1}{c}{\rotatebox[origin=l]{90}{\makecell[l]{shampoo\\bottle}}} & \multicolumn{1}{c}{\rotatebox[origin=l]{90}{\makecell[l]{standing\\bottle}}} & \multicolumn{1}{c}{\rotatebox[origin=l]{90}{tire}} & \multicolumn{1}{c}{\rotatebox[origin=l]{90}{valve}} & \multicolumn{1}{c}{\rotatebox[origin=l]{90}{wall}} & \multicolumn{1}{c}{\rotatebox[origin=l]{90}{\textbf{average}}} \\ 
			\midrule[0.8pt]
			FZ&	L&	
			87.96&	86.47&	73.85&	89.26&	74.87&	82.82&	90.89&	91.52&	93.19&	73.72&	84.23&	84.44	\\
			L&	L&	
			91.88&	89.30&	80.58&	89.91&	80.13&	91.44&	92.49&	91.94&	94.53&	77.33&	88.24&	87.98	\\			
			P&	L&	
			92.33&	87.78&	80.39&	89.95&	80.93&	90.30&	92.42&	92.07&	95.30&	77.94&	88.37&	87.98	\\
			\midrule[0.8pt]
			FZ&	FF&	
			90.84&	89.03&	75.99&	89.68&	78.42&	87.84&	91.75&	92.89&	93.66&	73.47&	87.90&	86.50	\\
			L&	FF&	
			92.08&	89.24&	80.79&	90.34&	81.14&	91.47&	92.65&	92.46&	94.12&	78.52&	90.09&	88.44	\\
			P&	FF&	
			92.25&	87.01&	79.78&	90.28&	80.49&	90.69&	92.93&	93.40&	94.85&	78.93&	89.02&	88.15	\\   
			\bottomrule[1.5pt]
			\end{tabular}
        \begin{tablenotes}
            \item FZ: Frozen; FF: Fully fine-tune; L: LoRA; P: Prompt layer. 
        \end{tablenotes}
        \end{threeparttable}
        \label{tab:finetune_SAM_results}
    \end{table}

Referring to Table~\ref{tab:finetune_SAM_results}, the results demonstrate a significant improvement compared to directly applying SAM to sonar images. Specifically, there is a minimum 44\% increase in the DICE score.
A comparison between the models employing a frozen Image encoder and those utilizing two different fine-tuning approaches reveals a 2-3\% improvement in DICE score. This discrepancy highlights the significance of fine-tuning the Image encoder to enhance performance when a domain gap exists between the source and target images. Furthermore, when examining the two fine-tuning methods for the Image encoder, no significant disparity in outcomes is observed (less than 0.3\% deviation in DICE score). Thus, both fine-tuning approaches demonstrate effectiveness. 
Upon comparing the performance of fine-tuning the Mask decoder with both LoRA and full-tuning, it is observed that the fully-tuned models exhibit a slightly superior performance overall. This discrepancy may be attributed to the fact that the lightweight nature of the Mask decoder facilitates its tuning process compared to the Image encoder. Furthermore, the process of full-tuning potentially enables a more effective adaptation to the target domain, thereby leading to improved fitting. 
Figure~\ref{fig:visual_results} (e) depicts the enhanced performance obtained by the fine-tuned SAM model with box prompts (Image encoder: L, Mask decoder: FF). 

\subsection{Fine-tuning SAM for semantic segmentation on sonar images}

Based on the provided information from the boxes prompts, the fine-tuned models outperform traditional deep-learning-based segmentation methods (e.g., FCN and DeeplabV3) in terms of DICE score. However, the use of box input limits its applicability to fully-automatic scenarios. Consequently, in this section, we assess the performance of SAM after fine-tuning it specifically for the semantic segmentation task. 
In our experiments, we assess four fine-tuning approaches for the Image encoder: frozen, fully-tuning, fine-tuning with LoRA, and fine-tuning with the Prompt layer. Additionally, we explore three variations of mask decoder settings: custom segmentation head, fully-tuning, and fine-tuning with LoRA. We use the vit-l backbone for the Image encoder. The outcomes of these evaluations are presented in Table~\ref{tab:semantic_SAM_results}.

\begin{table}[tbph]
    \footnotesize        
        \caption{Performance of different models fine-tuned on SAM for semantic segmentation evaluated by DICE Score.}
        \centering
		\setlength\tabcolsep{4.5pt}
        \begin{threeparttable}		
			\begin{tabular}{cccccccccccccc}			
			\toprule[1.5pt]	
			\multicolumn{1}{c}{\multirow{2}{*}{\rotatebox[origin=l]{90}{\textbf{\makecell[l]{Image\\encoder\quad\quad}}}}} & 
			\multicolumn{1}{c}{\multirow{2}{*}{\rotatebox[origin=l]{90}{\textbf{\makecell[l]{Mask\\decoder\quad\quad}}}}} & 
			\multicolumn{12}{c}{\textbf{DICE Score $(\%)$}} \\ 			
			\cline{3-14} 
			\specialrule{0em}{2pt}{2pt}
			\multicolumn{1}{c}{}	&\multicolumn{1}{c}{}                       & \multicolumn{1}{c}{\rotatebox[origin=l]{90}{bottle}} & \multicolumn{1}{c}{\rotatebox[origin=l]{90}{can}} & \multicolumn{1}{c}{\rotatebox[origin=l]{90}{chain}} & \multicolumn{1}{c}{\rotatebox[origin=l]{90}{\makecell[l]{drink\\carton}}} & \multicolumn{1}{c}{\rotatebox[origin=l]{90}{hook}} & \multicolumn{1}{c}{\rotatebox[origin=l]{90}{propeller}} & \multicolumn{1}{c}{\rotatebox[origin=l]{90}{\makecell[l]{shampoo\\bottle}}} & \multicolumn{1}{c}{\rotatebox[origin=l]{90}{\makecell[l]{standing\\bottle}}} & \multicolumn{1}{c}{\rotatebox[origin=l]{90}{tire}} & \multicolumn{1}{c}{\rotatebox[origin=l]{90}{valve}} & \multicolumn{1}{c}{\rotatebox[origin=l]{90}{wall}} & \multicolumn{1}{c}{\rotatebox[origin=l]{90}{\textbf{average}}} \\ 
			\midrule[0.8pt]	
			FF&	C&
			0.77&	0.02&	24.39&	0.04&	0.02&	0.01&	0.01&	0.01&	30.09&	0.04&	68.95&	11.31\\
			FZ&	C&
			75.08&	37.84&	62.72&	65.54&	7.94&	51.20&	66.31&	12.72&	76.64&	29.20&	81.83&	51.55\\
			L&	C&
			82.44&	62.05&	69.03&	82.05&	69.47&	76.82&	80.19&	89.38&	86.39&	63.44&	84.85&	76.92\\
			P&	C&	
			84.20&	67.15&	72.11&	81.22&	41.57&	72.23&	77.40&	82.61&	84.98&	59.66&	84.53&	73.42\\			
			\midrule[0.8pt]
			FZ&	FF&	
			73.75&	50.99&	58.29&	65.55&	19.19&	56.41&	73.51&	85.19&	72.14&	39.04&	81.16&	61.38\\
			L&	FF&	
			79.82&	66.26&	68.15&	84.07&	53.85&	77.65&	82.10&	90.30&	82.38&	54.60&	85.19&	74.94\\
			P&	FF&	
			88.01&	64.69&	67.22&	83.91&	60.86&	77.83&	82.31&	82.20&	84.33&	53.28&	85.11&	75.43\\
			
			\midrule[0.8pt]
			FZ&	L&	
			54.15&	23.19&	56.01&	50.24&	0.92&	18.37&	28.09&	10.25&	61.99&	27.44&	78.81&	37.22\\
			L&	L&	
			85.19&	59.90&	69.65&	80.75&	61.88&	76.22&	80.68&	86.89&	86.08&	68.40&	86.40&	76.55\\			
			P&	L&	
			84.73&	68.78&	67.97&	83.77&	62.33&	80.62&	80.05&	90.30&	89.19&	62.03&	85.37&	77.74\\
			\bottomrule[1.5pt]
			\end{tabular}
        \begin{tablenotes}
            \item C: Custom segmentation head; FZ: Frozen; FF: Fully fine-tune; L: LoRA; P: Prompt layer. 
        \end{tablenotes}
        \end{threeparttable}
        \label{tab:semantic_SAM_results}
    \end{table}

Based on Table~\ref{tab:semantic_SAM_results}, a noticeable decline in performance is observed when comparing models utilizing fine-tuned Image encoders with their frozen counterparts. These results align with the outcomes obtained in the previous section for fine-tuned models. Both adaptation methods employed on the Image encoder yield comparable results. However, fully tuning the Image encoder leads to significantly low performance (11.31\% on DICE score), possibly due to the limited availability of data for fine-tuning.
The performance of the models trained with a frozen Image encoder is similar when comparing different settings of the Mask decoder. Among these settings, the fully-tuned Mask decoder achieves the highest performance, while the Mask decoder fine-tuned with LoRA demonstrates the lowest performance. These findings align with the results presented in section~\ref{Fine_tuning_SAM}.
Nevertheless, the performance of models specifically optimized for semantic segmentation remains inferior compared to both traditional semantic segmentation networks~\cite{long2015fully,chen2018encoder} and models employing box prompts. This observation aligns with previous studies on fine-tuning SAM~\cite{MedSAM,qiu2023learnable,zhang2023customized}. The underlying reason for this discrepancy could be attributed to the limited utilization of the SAM network architecture in semantic segmentation tasks, resulting in a decline in performance. The outcomes are illustrated in Figure~\ref{fig:visual_results} (d).

\subsection{Impact of model scale on performance}
In general, larger backbone architectures tend to yield better performance for natural image tasks~\cite{kirillov2023segment,tan2019efficientnet}. However, during transfer learning, the performance can be influenced by the limited availability of samples in the target domain. In this section, we assess and compare the performance of fine-tuned models with varying backbone scales. Specifically, we investigate three variations of fine-tuning the Image encoder: freezing the Image encoder, fine-tuning the Image encoder with the prompt layer, and fine-tuning the Image encoder with LoRA. We present the experimental setup and results in Table~\ref{tab:different_scales}. The item attaining superior performance within each comparison group is emphasized using underlining.

\begin{table}[tbph]
    \footnotesize        
	\caption{Performance of the models with different scale backbones evaluated by DICE Score.}
	\centering
	\begin{threeparttable}		
		\begin{tabular}{cccc}			
		\toprule[1.5pt]
		\textbf{Backbone} & \textbf{Image encoder} & \textbf{Mask decoder} & \textbf{Avg. DICE Score $(\%)$} \\ 		
		\midrule[0.8pt]
		vit-b&	Frozen&	Custom segmentation head&	42.51\\
		vit-l&	Frozen&	Custom segmentation head&	51.55\\
		vit-h&	Frozen&	Custom segmentation head&	\underline{52.77}\\
		\midrule[0.8pt]
		vit-b&	Prompt-layer&	Custom segmentation head&	67.13\\
		vit-l&	Prompt-layer&	Custom segmentation head&	\underline{73.42}\\
		vit-h&	Prompt-layer&	Custom segmentation head&	72.72\\
		\midrule[0.8pt]
		vit-b&	LoRA&	Custom segmentation head&	73.74\\
		vit-l&	LoRA&	Custom segmentation head&	76.92\\
		vit-h&	LoRA&	Custom segmentation head&	\underline{77.05}\\
		\bottomrule[1.5pt]
		\end{tabular}
	\end{threeparttable}
	\label{tab:different_scales}
\end{table}

Table~\ref{tab:different_scales} illustrates the relationship between model backbone scale and performance. The smallest backbone, vit-b, exhibits the lowest performance. In contrast, models with larger backbones (vit-l and vit-h) demonstrate improved performance, with average DICE scores increasing by approximately 9\%, 6\%, and 3\%, respectively, compared to the vit-b version. However, we observe no significant performance gain when transitioning from the backbone vit-l to vit-h. This suggests a potential performance saturation due to limited training samples in the target domain, as larger parameter counts generally require more training data. These results indicate that selecting an appropriate scale backbone for a specific task is crucial for achieving computationally effective outcomes.

\section{Conclusion}

In this study, we propose the integration of SAM into sonar imaging scenarios, specifically focusing on fine-tuning SAM for sonar image analysis. Directly applying SAM to sonar images poses a challenge due to the prominent domain gap between the training samples used in SAM. To address this issue, we employ various fine-tuning methods to adapt SAM to sonar imaging. Through the fine-tuning process, SAM models exhibit substantial improvements in performance. Additionally, we explore the fine-tuning of SAM for semantic segmentation tasks and assess the impact of backbone scale on segmentation performance. Our experiments demonstrate the potential of SAM, a large vision model, in sonar imaging scenarios and provide valuable insights for further research.

\section*{Acknowledgments}
This work was supported by the National Natural Science Foundation of China [grant number 42276187]; and the Fundamental Research Funds for the Central Universities, China [grant number 3072022FSC0401].

\bibliography{mybibfile}

\end{document}